\documentclass[11pt]{article}

\usepackage[preprint]{acl}

\usepackage{times}
\usepackage{latexsym}

\usepackage[T1]{fontenc}

\usepackage[utf8]{inputenc}

\usepackage{microtype}

\usepackage{inconsolata}

\usepackage{graphicx}

\usepackage{booktabs}
\usepackage{amsmath,amssymb}
\usepackage{enumitem}
\usepackage{tikz}
\usetikzlibrary{arrows.meta,positioning}
\usepackage{xcolor}

\colorlet{promptcolor}{green!8}
\colorlet{ctxcolor}{blue!8}
\colorlet{freecolor}{gray!15}
\colorlet{tgtcolor}{red!8}

\newcommand{\plvr}{P-LVR}
\newcommand{\nlvr}{N-LVR}
\newcommand{\dlvr}{D-LVR}

\newcommand{\prism}{\textsc{Prism}}
\newcommand{\vbench}{V$^{*}$Bench}

\title{Cosine Misleads: Auxiliary Losses Reshape Vision Language Models, Not Their Latents}

\author{XiuYu Zhang \and Junfeng Fang \and Zhenkai Liang \\
        National University of Singapore}

\begin{document}
\maketitle
\begin{abstract}
Latent visual reasoning (LVR) inserts supervised latent tokens between perception and answer generation in vision-language models (VLMs). The field uses alignment between these latents and their visual targets, i.e., cosine similarity or mean squared error (MSE), as both the training loss and the quality metric, assuming that better alignment yields a better answer. We test this with a designed matrix of five LVR variants and find the assumption inverted: cosine alignment is negatively correlated with accuracy across all five (r=-0.94). To explain this, we introduce PRISM, a pair of inference-time diagnostics: a linear probe that asks where the answer is decodable, and a corruption test that asks whether the latent is load-bearing. The supervised latents are largely bypassed. Corrupting them shifts accuracy by at most four points. The answer is decodable downstream of the latent but not at it, and the size of this decodability gap predicts how much each variant relies on its latent under perturbation. Consistent with an Information Bottleneck reading of the loss, the auxiliary objective reshapes the language model via shared parameters rather than via the latent variable it nominally optimizes.  
\end{abstract}

\section{Introduction}

Latent visual reasoning (LVR) inserts continuous-value latent tokens between visual perception and answer generation in a vision-language model (VLM)~\citep{bordes2024introductionvisionlanguagemodeling}. These latents are supervised against teacher-forced visual targets during training and fed back autoregressively at inference. 
In this active research area~\citep{li2026latent, yang2025machine, dong2025interleaved, monet2025latent, valr2026}, the alignment signals, e.g., mean square error (MSE), cosine similarity, and their variant, between the generated latents and target representations, have been used as both the training loss and the post-hoc quality metric. 
The implicit assumption behind this practice is that a better cosine means a more faithful latent, which implies a better answer.

\begin{figure}[!t]
  \centering
  \includegraphics[width=\linewidth]{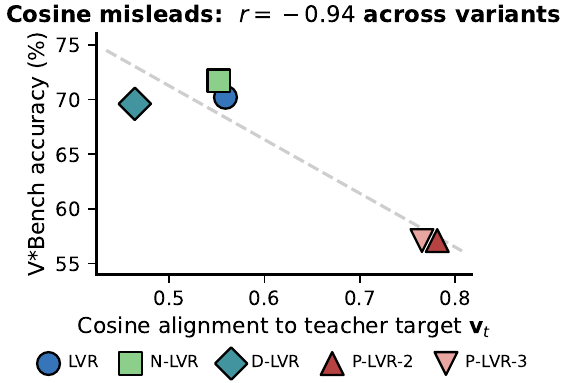}
  \caption{\textbf{Cosine misleads.} Cosine alignment between LVR-position hidden states and their teacher-forced visual targets is negatively correlated with V$^{*}$Bench accuracy across all five trained variants (Pearson $r{=}{-}0.94$). Progressive variants (P-LVR-2, P-LVR-3) reach the highest cosine but the lowest accuracy.}
  \label{fig:teaser}
\end{figure}

To verify what cosine alignment actually indicates, we designed five LVR variants that vary in how the latents are trained (from reconstruction-pure to input-noise-regulated to progressively scaffolded) and trained each on a shared backbone, data, and step budget.
Surprisingly, cosine similarity is negatively correlated with \vbench{}~\citep{wu2024vstar} accuracy across all five (Pearson $r=-0.94$). 
A progressive scaffolding variant raises cosine alignment by $40\%$ over the baseline and lowers reconstruction error from $3.79$ to $1.55$, but loses 13 \vbench{} points. This decrease replicates on BLINK~\citep{fu2024blink} and MMVP~\citep{tong2024eyes}.
A noise-train variant matches the baseline's cosine alignment to three decimals (0.556 vs. 0.555), but gains 1.5 \vbench{} points and responds in the opposite direction when its latents are zeroed during inference, i.e., its own latent helps while the baseline's hurts. 
Cosine similarity is thus shown to be a misleading signal in isolation.

To investigate, we apply two diagnostics that we collectively call \prism{}: a corruption test on the latents and a linear probe of the answer-decoding hidden state.
First, the corruption test asks whether the model actually uses the latents at inference.
We intervened in the LVR positions at each generation step by truncating them to zero, perturbing them with Gaussian noise, or replacing them with latents from another sample. 
Across all five variants, every intervention shifts \vbench{} by at most four points. For the worst variant, zeroing the latents improves accuracy.  
As a result, the LVR latents -- the very tokens cosine optimizes -- are largely bypassed at inference. 

However, the same five variants differ by 13 \vbench{} points. If latents are bypassed, the training objective must do something else, not through the latents themselves directly. 
\prism{}'s second diagnostic answers this. 
We fit a linear probe at two positions: the answer-decoding state (the hidden state the LM head reads to produce the answer), and the feedback variable (the latent the autoregressive loop re-injects).
The answer is decodable from the answer-decoding state, which is expected. The interesting question is whether it is also decodable from the feedback variable: it is not.
The signal sits downstream of the latent, rather than in the latent, where the loss is optimized.
The contrast between probe accuracy at the two positions varies substantially across the matrix, and the size of this contrast predicts how each variant responds to latent perturbation.

Under an Information Bottleneck~\citep[IB]{tishby2000informationbottleneckmethod} reading, this can be explained as the dominating cross-entropy (CE) term in the training loss applies relevance pressure to whatever computation produces the answer, not to the supervised latents specifically. 
Therefore, the model is free to route the answer without using the latents under the current LVR loss construction. 
The training objective reshapes the language model through gradient flow into shared parameters, not through what the latent encodes. 
The auxiliary loss works, but not for the reason its name suggests.

We hypothesize this pattern is not specific to LVR. 
Auxiliary losses across multimodal learning supervise intermediate representations against external targets, and the assumption that the supervised representation is also the load-bearing one is undertested. 
\prism{} operationalizes a test: the probe localizes where the answer is decodable; the corruption confirms whether the latents are load-bearing.
Together, they make visible what cosine cannot see.

Our contributions can be summarized as follows:
\begin{enumerate}[nosep,leftmargin=*]
\item \textbf{A designed test of the LVR training assumption.} 
Across a matrix of five variants spanning the design space for supervising latents, cosine ranks variants approximately backwards relative to accuracy, and the latents the loss optimizes are largely bypassed at inference. 

\item \textbf{The training objective reshapes the 
VLM through shared parameters.} 
Despite the bypass, variants differ substantially in accuracy. The answer-relevant signal is linearly decodable at the model's answer-decoding state, but not in the latent loop re-injects. The size of this contrast predicts both task accuracy and latent reliance under perturbation.

\item \textbf{\prism{} as a replacement diagnostic.} 
Linear probes at two positions reveal a contrast that locates where the answer sits in the model, and a corruption test checks whether the supervised latent is load-bearing. The two are empirically connected as the size of the probe contrast predicts the corruption response.

\end{enumerate}
\section{Related Work}

\paragraph{Latent visual reasoning and the alignment assumption.}
LVR~\citep{li2026latent} generates latent tokens supervised to reconstruct visual embeddings from task-relevant regions defined by bounding boxes via MSE. 
Concurrent and subsequent works vary in their supervision to align latent representations and visual targets: Mirage~\citep{yang2025machine} uses cosine as the loss, Monet~\citep{monet2025latent} uses cosine to construct the loss, VaLR~\citep{valr2026} adopts the REPA~\citep{Leng_2025_ICCV} loss, which uses cosine to measure similarity.  
Across these works, teacher-feature alignment is treated as both a training signal and a post-hoc quality metric. 
This dual use is rarely tested directly or receives the attention it deserves. 
Inspired by this trend, we construct a matrix of LVR variants that span its design choices, serving as a testbed to directly test the alignment assumption.

\paragraph{Probing and faithfulness.}
Linear probing trains a single linear classifier from frozen hidden states to predict a target label~\citep{alain2017understanding}. 
It is used to study and understand the internal states of neural networks. 
Subsequent work showed that probe accuracy can reflect probe capability rather than property presence~\citep{hewitt-liang-2019-designing, belinkov-2022-probing}, which motivates control-task selectivity.
Faithfulness work in chain-of-thought (CoT) reasoning in large language models (LLMs) shows that explanations can be unfaithful to the model's actual computation~\citep{turpin2023language}, so causal reliance is better tested by intervening on the reasoning traces~\citep{tutek-etal-2025-measuring}. 
Drawing lessons from these works, we applied the two ideas in \prism{} to analyze a latent-reasoning setting. 
We use a probe to ask whether the answer is decodable from a given hidden state, and a corruption test to ask whether that state is causally load-bearing.

\paragraph{Information bottleneck.}
The Information Bottleneck~\citep[IB]{tishby2000informationbottleneckmethod} frames representation learning as a trade-off between compressing input and preserving information about a target. 
The variational form~\citep{alemi2017deep} makes this tractable. 
We use IB as interpretive scaffolding for our results, since cosine and MSE optimize only one side of the IB objectives, leaving the other to implicit pressure from the next-token prediction loss.
This may be one way to explain why the cross-variant correlation between cosine and accuracy is free to run in any direction. 
We use IB as interpretive scaffolding and do not claim IB or its usage as a contribution.
\section{The Information Bottleneck view of LVR}
\label{sec:ib}

This section introduces the lens through which we interpret the empirically observed cosine dissociations in Section~\ref{sec:expt}. 
We frame the LVR latents as a representation in the Information Bottleneck (IB) sense and argue that the reconstruction-based loss (such as MSE and cosine) the field uses does not bound either side of the IB objective.

\subsection{The LVR loss}

Let $\mathbf{h}_{t}$ be the LVR-position hidden state at iteration $t$, and $\mathbf{v}_t$ be the teacher-forced visual target for that position. 
LVR~\citep{li2026latent} supervises the latents with a reconstruction loss alongside the typical next-token prediction loss:
\begin{equation}
\label{eq:lvr-loss}
\mathcal{L}_{\text{LVR}} = \mathcal{L}_{\text{CE}} + \lambda \cdot \text{MSE}(\mathbf{h}_{t}, \mathbf{v}_t).
\end{equation}

Cosine-based variants~\citep{yang2025machine, monet2025latent, valr2026} replace the MSE term with cosine or patch-wise alignment losses or mix with other losses. 
The form differs, but the role is the same, i.e., pull $\mathbf{h}_{t}$ towards $\mathbf{v}_t$.
In inference, the model produces its own latents and feeds them back autoregressively.

\begin{figure*}[t]
\centering
\includegraphics[width=\textwidth]{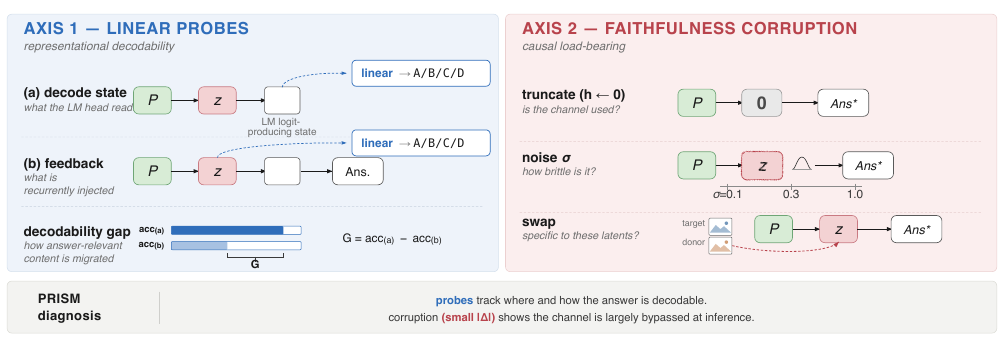}
\caption{\textbf{\prism{} overview.} Two inference-time diagnostics for the LVR family. \textbf{Axis~1} trains linear probes~\citep{alain2017understanding} at two positions in the model: (a) the answer-decoding state the LM head reads, and (b) the feedback variable the autoregressive loop re-injects. We report the \emph{decodability gap} $G = acc_{probe}(a) - acc_{probe}(b)$, which summarizes how much more answer-decodable the post-latent state is than the latent. \textbf{Axis~2} perturbs the LVR-injected hidden states (latents) in generation (truncation, noise, swap) and measures the change in accuracy; small $|\Delta\text{acc}|$ means the latent is bypassed. Section~\ref{sec:expt} shows the two axes are tied: $G$ predicts each variant's response to latent perturbation.}
\label{fig:prism}
\end{figure*}

\subsection{The IB objective}

Let $X$ be the input (image, question, prefill context) and $Y$ be the answer. The IB writes the latent-quality trade-off as a Lagrangian:

\begin{equation}
\label{eq:ib-lagrangian}
\mathcal{L}_{\text{IB}} = \beta\, I(X; Z) - I(Z; Y),
\end{equation}
where $Z$ is the intermediate representation, $I(\cdot;\cdot)$ is mutual information, and $\beta>0$ controls compression. 
An IB-optimal $Z$ is compressed (small $I(X; Z)$) while preserving information about the answer (large $I(Z; Y)$).

Three intermediate representations are used in our LVR setting. 
We will use $Z_{ch}$ for the LVR-position hidden state itself, i.e., the object the loss is computed on, and the IB objectives are about. 
We will use $Z_{fb}$ for the feedback variable re-injected into the VLM at the next position. 
Without a projection, $Z_{ch}=Z_{fb}$. 
We will use $H_{ans}$ to denote the answer-decoding hidden state read by the LM head at the first ordinary text token. 
Section~\ref{sec:expt} probes $Z_{fb}$ and $H_{ans}$ separately following \prism{}. 

\subsection{The LVR loss is an indirect IB objective, and cosine measures only one side of it}

The LVR loss (Equation~\ref{eq:lvr-loss}) has two terms, and each does part of what IB asks for. 
The MSE term serves as a proxy for compression. $\mathbf{v}_t$ is the visual embedding of an annotated region, i.e., a hand-selected, low-rate, task-relevant representation of $X$. 
Driving $\text{MSE}(\mathbf{h}_{t}, \mathbf{v}_t)$ towards zero pulls $Z_{ch}$ to match $\mathbf{v}_t$ in every dimension. 
In the limit, $Z_{ch}=\mathbf{v}_t$ and thus $I(X; Z_{ch})$ inherits $\mathbf{v}_t$'s small rate. 
The loss does not directly bound $I(X;Z)$, but it implements compression indirectly by anchoring $Z_{ch}$ to a target whose rate is low by construction. 
Cosine alignment is a looser version of the same proxy: it pins direction but not magnitude or orthogonal components, so the inheritance chain is weaker, but the basic mechanism remains the same.

The cross-entropy (CE) term supplies relevance, but through a different route. 
CE on $Y$ applies pressure on whatever computation produces the answer.
Since $Z_{ch}$ sits inside that computation path during training, CE gradients flow back through it and shape its content.
Critically, the pressure is on the system as a whole, not on $Z_{ch}$ in particular. 
The model is free to satisfy CE by making $Y$ depend on whichever parts of its state are convenient, and nothing in the loss requires that $Z_{ch}$ be among those parts.

This is the source of the dissociations we report. 
Cosine measures how well the MSE proxy performs, i.e., how close $Z_{ch}$ is to $\mathbf{v}_t$ and how effectively the proxy compression succeeds. 
It does not measure where relevance has settled.
Section~\ref{sec:expt} shows that at inference, removing $Z_{ch}$ entirely shifts accuracy by at most four points, which means that the model has routed relevance through other parts rather than the latents. 
The CE pressure that ran through $Z_{ch}$ during training has been satisfied elsewhere.

Therefore, cosine reports on one half of an indirect IB objective and is silent on the other half. 
A high cosine tells that the latent is well-anchored to its target.
It does not indicate whether the latent carries any load, i.e., whether the model actually uses the latent to produce the answer.
In Section~\ref{sec:prism}, we propose a measurement that asks that question: the decodability gap between the answer-decoding state and the feedback variable (last latent). 
The Section~\ref{sec:expt} cross-variant pattern is what that looks like empirically: cosine moves across a 40\% range, and accuracy moves independently because the two are answering different questions under the current LVR loss.

\section{\prism{}: a replacement diagnostic for cosine}
\label{sec:prism}

Cosine measures the fidelity of a latent to its supervision target. 
The findings in Section~\ref{sec:expt} show that this is a misleading question for analyzing the LVR family: the supervised latents are largely bypassed at inference, so target fidelity tells little about where the answer actually lies.
\prism{} replaces cosine with two inference-time diagnostics: a linear probe that asks where the answer is decodable, and a corruption test that asks whether the supervised latent is load-bearing. 

\begin{figure*}[!t]
\centering
\includegraphics[width=\textwidth]{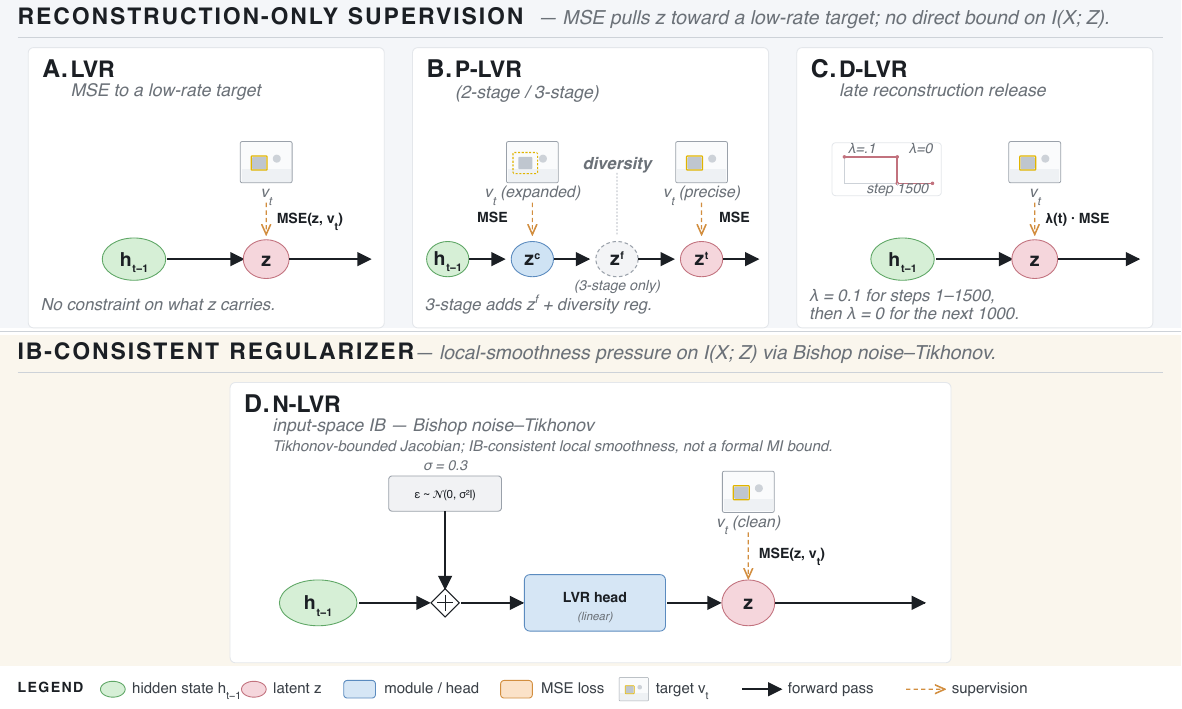}
\caption{\textbf{The matrix of five LVR variants, grouped by their relationship to the IB objective.} Top row (reconstruction-only supervision): LVR baseline, \plvr{} (progressive 2/3-stage scaffolding), \dlvr{} (reconstruction loss removed at mid-training). All three rely on an MSE-to-low-rate target as an indirect compression mechanism, with no direct bound on $I(X;Z)$. Bottom (IB-consistent regularizer): \nlvr{} adds zero-mean Gaussian noise to the teacher-forced input during training while keeping the target clean, which to first order is a Tikhonov-style Jacobian penalty on the encoder, i.e., a local-smoothness prior approximating IB-style suppression of channel rate.}
\label{fig:architecture}
\end{figure*}

\subsection{Axis 1: linear probes}
Where the answer is decodable tells where the loss actually occurred. 
To find out, for any LVR model and any multiple-choice question, we run the model once in the answer-decoding mode and extract two hidden-state vectors. 
Position (a) is the answer-decoding states: the hidden state at the iteration that produces the answer-text token, after the model has consumed the LVR tokens. 
This is the state the LM head reads to predict the answer.
Position (b) is the feedback variable: the hidden state at the LVR boundary that the autoregressive loop re-injects as the next-position input embedding.

We fit logistic regression at each position $p\in \{a,b\}$ from the hidden state to the answer letter using 5-fold class-stratified cross-validation, $\ell_2$ regularization, and per-fold input standardization~\citep{alain2017understanding}.
We denote by $acc_{probe}$ the resulting cross-validated accuracy.
The Barber--Agakov variational bound \citep{barber2003information} gives
\begin{align*}
&I(\mathrm{rep};Y) \\
&\ge 
H(Y) + \mathbb{E}_{(\mathrm{rep},Y)\sim p_{\mathrm{data}}}
\left[\log q(Y\mid \mathrm{rep})\right] \\
&=
H(Y)-\mathrm{CE}(q),
\end{align*}
where $H(Y)$ denotes the entropy of the answer variable $Y$, $p_{\mathrm{data}}(\mathrm{rep},Y)$ is the empirical joint distribution induced by the held-out examples, $q(Y\mid \mathrm{rep})$ is a variational predictor of the answer from the representation.
Thus, a low held-out cross-entropy provides a lower bound on how much answer-relevant information the representation carries.

The two probe accuracies describe a contrast: how much more decodable the answer is at the model answer-decoding state than at the latent loop hands forward.
In any working model, $acc_{probe}(a)$ will be high as position (a) is the state the LM head reads to produce an answer. 
The empirical question is what happens at (b): does the latent loop hand forward carry the answer, or has the answer signal settled elsewhere?
We summarize the contrast with the decodability gap $G\equiv acc_{probe}(a) - acc_{probe}(b)$, but report both probes throughout, because the gap and its components carry different information.
Section~\ref{sec:expt} reports all three across the variant matrix, and the connection to Axis 2.

\begin{table*}[!t]
\centering
\small
\setlength{\tabcolsep}{4pt}
\caption{\textbf{Per-variant measurements across the LVR matrix.} \textbf{cos}: mean cosine alignment between LVR-position hidden state and target visual embedding. \textbf{V$^{*}$B / MMVP / BLINK}: benchmark accuracy (\%). \textbf{Acc(a) / Acc(b)}: linear-probe accuracy (\%) at the answer-decoding state $H_{\text{ans}}$ and the feedback variable $Z_{\text{fb}}$, under 5-fold class-stratified cross-validation. \textbf{G}: probe contrast (decodability gap), $\text{Acc}(a)-\text{Acc}(b)$ in percentage points. \textbf{trunc / $\sigma$/ swap}: faithfulness-corruption $\Delta$accuracy on V$^{*}$Bench (signed; negative means corruption \emph{hurts}, i.e.\ own latents help).}
\label{tab:probe-gap-pervariant}
\begin{tabular}{l c c c c c c c c c c c c}
\toprule
 & \textbf{cos} & \multicolumn{3}{c}{\textbf{Accuracy}} & \multicolumn{3}{c}{\textbf{Probes}} & \multicolumn{5}{c}{\textbf{Faithfulness corruption $\Delta$acc}} \\
\cmidrule(lr){2-2} \cmidrule(lr){3-5} \cmidrule(lr){6-8} \cmidrule(lr){9-13}
\textbf{Variant} & & \textbf{V$^{*}$B} & \textbf{MMVP} & \textbf{BLINK} & \textbf{(a)} & \textbf{(b)} & \textbf{G} & \textbf{trunc} & \textbf{$\sigma$=.1} & \textbf{$\sigma$=.3} & \textbf{$\sigma$=1.0} & \textbf{swap} \\
\midrule
LVR     & 0.555 & 70.2 & 49.7 & 53.4 & 69.1 & 32.5 & 36.6 & $-2.6$ & $-1.6$ & $+1.0$ & $-1.0$ & $+1.6$ \\
N-LVR   & 0.556 & 71.7 & 50.0 & 52.9 & 66.0 & 41.9 & 24.1 & $-2.6$ & $-0.5$ & $-1.0$ & $-2.1$ & $-0.5$ \\
D-LVR   & 0.464 & 69.6 & 49.3 & 51.4 & 64.9 & 33.0 & 31.9 & $-1.1$ & $\phantom{+}0.0$ & $+0.5$ & $\phantom{+}0.0$ & $+2.1$ \\
P-LVR-2 & 0.777 & 57.1 & 48.7 & 47.3 & 50.2 & 35.6 & 14.6 & $\phantom{+}0.0$ & $+2.1$ & $\phantom{+}0.0$ & $-0.5$ & $+2.1$ \\
P-LVR-3 & 0.769 & 57.1 & 48.0 & 48.5 & 48.7 & 34.6 & 14.1 & $+2.1$ & $+2.6$ & $+4.2$ & $+3.1$ & $+2.6$ \\
\bottomrule
\end{tabular}
\end{table*}

\subsection{Axis 2: faithfulness corruption}

We perturb the LVR latents at each generation and measure the change in benchmark accuracy relative to a clean pass. 
The three perturbations are truncation ($h\leftarrow 0$), additive Gaussian noise at $\sigma \in \{0.1, 0.3, 1.0\}$, and random-donor swap ($h\leftarrow h_{donor}$, where $h_{donor}$ is another sample's clean latent at the matching iteration). 
The corruption propagates into the KV cache at every layer; for K/V projections without bias, truncation yields literally zero K and V at the corrupted position.

A small change in accuracy ($|\Delta\text{acc}|$) under intervention is what bypass looks like operationally, i.e., the model does not rely on the latent under that intervention. 
The sign of $\Delta$ separates the case where the latent helps from the case where it hurts: own latents better than random against own latents worse. 
The three perturbations probe different things.
Truncation asks the binary question, i.e., is the latent used at all? 
Noise grades robustness as a function of perturbation magnitude. 
Swap asks whether the answer is content-specific to this sample's latent or if the model would tolerate a random donor. 
\section{Experiments}
\label{sec:expt}

\begin{table}[!t]
\centering
\small
\setlength{\tabcolsep}{4pt}
\caption{\textbf{Cross-variant Pearson correlations} of each diagnostic against \vbench{} accuracy, the truncation and small-noise ($\sigma=0.1$) corruption response.}
\label{tab:probe-gap-corr}
\begin{tabular}{l c c c}
\toprule
\textbf{Diagnostic} & \textbf{vs V$^{*}$B} & \textbf{vs trunc $\Delta$} & \textbf{vs $\sigma$=.1 $\Delta$} \\
\midrule
cosine       & $-0.94$ & $+0.75$ & $+0.84$ \\
probe-(a)    & $+0.98$ & $-0.92$ & $-0.98$ \\
probe-(b)    & $+0.20$ & $-0.26$ & $-0.02$ \\
$G$          & $+0.86$ & $-0.77$ & $-0.93$ \\
\bottomrule
\end{tabular}
\end{table}

\subsection{The LVR variant matrix and setup}

We designed and trained five variants of the LVR loss on a shared backbone, data, and step budget. 
The variants were chosen to span the common design choices the family uses to train its latent tokens. %
The shared loss is $\mathcal{L}_{\text{LVR}} = \mathcal{L}_{\text{CE}} + \lambda \cdot \text{MSE}(\mathbf{h}_{t}, \mathbf{v}_t)$ with $\lambda=0.1$ throughout.
Each variant modifies how supervision is applied, which target it uses, or how latent positions are organized.

\textbf{LVR}~\citep{li2026latent} is the unconstrained baseline: a single latent block of $K$ positions, each supervised against its teacher visual embedding, with no additional constraints.
\textbf{N-LVR} adds zero-mean Gaussian noise ($\sigma=0.3$) to the teacher-forced input during training while keeping the target clean. 
To first order, this is a Tikhonov-style Jacobian penalty on the encoder~\citep{bishop1995training}, i.e., a local-smoothness prior that approximates IB-style suppression of channel rate.
\textbf{D-LVR} resumes from the LVR checkpoint at step 1500 and continues for 1000 more steps with $\lambda_{rec}=0$, testing what the VLM does when reconstruction pressure is removed mid-training while CE pressure continues.
\textbf{P-LVR-2} and \textbf{P-LVR-3} split the latent block into stages: context $\rightarrow$ target (P-LVR-2) or context $\rightarrow$ free $\rightarrow$ target (P-LVR-3).
In which each stage (except the free stage) is teacher-forced against its own bounding box embedding, with the context box expanded by $\alpha=1.5$.

The backbone is Qwen2.5-VL-3B-Instruct~\citep{bai2025qwen25vltechnicalreport}, with the vision tower and visual merger frozen and the language model trainable. 
The training data is Visual-CoT~\citep{shao2024visual}: 438k question-answer pairs with task-relevant bounding boxes.
All variants train for 2500 steps with a learning rate of $10^{-5}$ on a cosine schedule, in bf16 precision, with an effective batch size of 64. 
We evaluate on \vbench{}~\cite{wu2024vstar} for visual search in cluttered scenes, MMVP~\citep{tong2024eyes} for paired-option comparison, and BLINK~\citep{fu2024blink} for perception-oriented multiple choice across five validation subsets.

\subsection{Cosine inverts the ordering it claims to measure}

Cosine alignment does not just fail to predict accuracy -- across the matrix, it predicts accuracy backward.
Table~\ref{tab:probe-gap-pervariant} reports cosine alignment to teacher targets and \vbench{} accuracy across the five variants. 
The cross-variant Pearson correlation is $r=-0.94$ (Table~\ref{tab:probe-gap-corr}), i.e., cosine and accuracy move in opposite directions. 

Beyond the surprising correlation, two specific dissociations pin down the kind of failure cosine exhibits. 
The first is directional.
The progressive scaffolding variants raise cosine from the baseline's 0.555 to 0.777 and 0.769 (a 40\% gain) and simultaneously lose 13 \vbench{} points. 
The progressive design trades training simplicity for cleaner reconstruction at the teaching stage. 
Cosine rewards that trade and \vbench{} punishes it.
The second dissociation is about resolution. LVR and \nlvr{} have nearly identical cosine values (0.555 vs.\ 0.556), but different \vbench{} accuracies and reverse the sign of their swap response: LVR's own latents hurt (+1.6), N-LVR's help (-0.5).
Cosine cannot distinguish between these two variants, even though every other measurement in the paper does.

A third dissociation surfaces once corruption is reported (Table~\ref{tab:probe-gap-corr}). 
Cosine correlates positively with corruption response: $r=+0.75$ against truncation $\Delta$ and $r=+0.84$ against small-noise ($\sigma=0.1$).
Variants with higher cosine are more helped by perturbing their latents, and the LVR baseline (which has middling cosine) shows the largest harm from truncation ($\Delta=-2.6$).
As a result, cosine fails to predict accuracy and points in the wrong direction regarding whether the latent is useful to the model. 

\begin{figure}[!t]
  \centering
  \includegraphics[width=\linewidth]{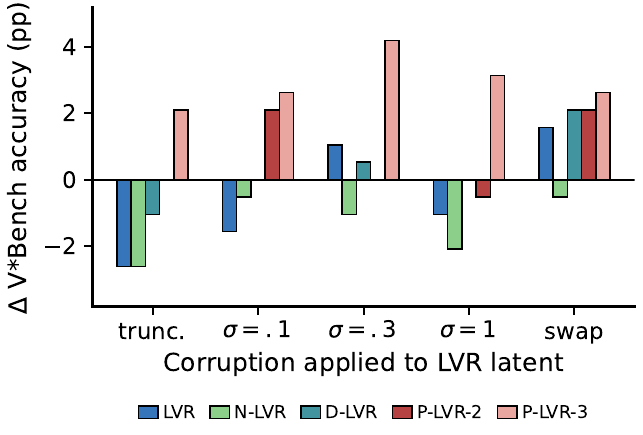}
  \caption{\textbf{Faithfulness corruption profiles on V$^{*}$Bench.} Change in \vbench{} accuracy ($\Delta\,\text{acc}$, pp) under each corruption applied to the LVR latents at latent reasoning steps, against a clean pass. Negative $\Delta=$ own latent helps; positive = own latent hurts.}
  \label{fig:faithfulness}
\end{figure}

\begin{figure*}[t]
\centering
\includegraphics[width=\textwidth]{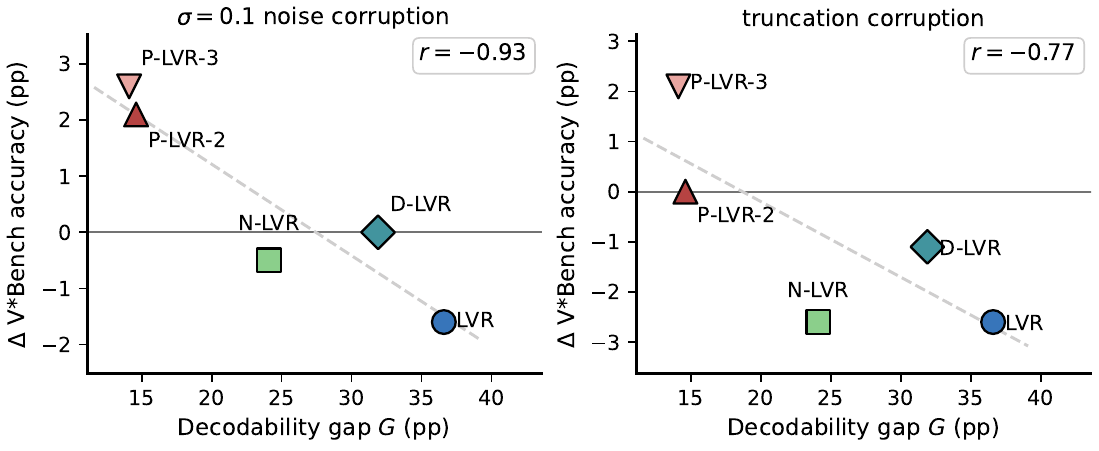}
\caption{\textbf{The probe contrast predicts latent reliance.} Decoding gap $G$ against corruption response under small-noise ($\sigma=0.1$, left) and truncation (right), across the five variants.}
\label{fig:decodability-gap}
\end{figure*}

\subsection{The latents are largely bypassed at inference}

We applied PRISM's Axis 2 corruption test at every LVR latent reasoning step: zeroing the latent hidden state (truncation), perturbing it with Gaussian noise at $\sigma \in \{0.1, 0.3, 1.0\}$, or replacing it with another sample's latents at the matching iteration (random-donor swap).
Figure~\ref{fig:faithfulness} shows the resulting change in \vbench{} accuracy per variant per corruption.

Across all five variants, every intervention shifts \vbench{} by at most four points. 
For P-LVR-3 with high reconstruction fidelity, zeroing the latents actually improves accuracy, suggesting that the latents were actively interfering. 
N-LVR is the only variant where own latents reliably beat random under swap, and for the rest, own latents are mildly worse than what a random donor provides.

This rules out a natural reading of the cross-variant \vbench{} spread.
The 13-point spread cannot be explained primarily by what the latent encodes, because the latents are bypassed by similar margins throughout the matrix.
Whatever distinguishes variants from one another lives in the model the loss has shaped, not in the latent content the loss optimizes. 
The corruption response is also informative in its own right. 
It differentiates variants along a dimension that cosine cannot see, i.e., LVR and N-LVR have nearly identical cosines but opposite signs of the swap $\Delta$.

\subsection{The probe contrast localizes where the answer lives}

Cosine and corruption agree on the negative finding that the latents are not where the answer lives, but neither tells where it has gone.
\prism{}'s Axis 1 does. For each variant, we fit a linear probe at the answer-decoding state, with accuracy $acc_{probe}(a)$, and at the feedback variable, with accuracy $acc_{probe}(b)$.
The two accuracies and their gap $G=acc_{probe}(a)-acc_{probe}(b)$ are reported in Table~\ref{tab:probe-gap-pervariant}.

$acc_{probe}(a)$ ranges from 50.2 (P-LVR-3) to 69.1 (LVR), and tracks \vbench{} across the matrix at $r=+0.98$ (Table~\ref{tab:probe-gap-corr}). 
$acc_{probe}(b)$ is much lower, between 32.5 and 41.9, and does not track \vbench{} ($r=+0.20$).
The contrast between the two ($G$) ranges from 14.1 to 36.6 percentage points, and is itself a strong predictor of \vbench{} ($r=+0.86)$.
The unconstrained baseline shows the largest gap, while the progressive variants show the smallest. 

All three quantities point at the same conclusion. 
The answer-relevant signal sits downstream of the supervised latents, not in the latent content the loop re-injects.
Probe accuracy at the model's answer-decoding state varies across variants in step with task accuracy.
Probe accuracy at the feedback variable varies independently of it.
Cosine reports on the latent's fidelity to its target, but the latent does not carry the answer.
The probe contrast reports the gap between latent and downstream decoding, and that gap tracks the model's competence on the task.

\subsection{The probe contrast predicts latent reliance}

Axis 1 measures where the answer is decodable.
Axis 2 assesses whether the model causally uses the latents.
These are different questions, but on this matrix they are empirically tied (Table~\ref{tab:probe-gap-corr}). 
$acc_{probe}(a)$ correlates with the corruption response under truncation at $r=-0.92$ and under small-noise ($\sigma=0.1$) at $r=-0.98$. 
The gap $G$ shows the same pattern at $r=-0.77$ and $r=-0.93$ (Figure~\ref{fig:decodability-gap}). 
Variants where the answer-decoding state is more answer-informative also rely on their latent more under perturbation.

As Figure~\ref{fig:decodability-gap} shows, when the decodability gap is large, i.e., when the answer has settled at the decoding state but not at the latent, the latent still plays a structural role in the path that produces the answer, and removing or lightly perturbing it disrupts downstream computation.
When the contrast is small (the P-LVR variants), the latent and downstream states are close to each other, and truncating the latent does not disrupt anything specific.
Bypass is the average behavior across the matrix.
The probe contrast predicts where each variant sits on the bypass spectrum.

This connection is the strongest empirical statement that \prism{} enables, and it requires both axes. 
Cosine cannot see the probe contrast. 
The corruption test alone cannot determine where the answer lies. 
The probe alone cannot see what the model actually uses. 
The connection between the probe contrast and corruption response is visible only when both diagnostics are applied to the same matrix.
\section{Discussion}
\label{sec:discussion}

Recent LVR works were built on the assumption that a latent supervised to match a visual target will carry that visual information into the answer. 
Our results show the assumption can fail without the method failing: the supervised latents are largely bypassed, yet the variants still differ by 13 \vbench{} points because the auxiliary loss reshapes the VLM through gradient flow into shared parameters.
The generalizable point is that an auxiliary loss can act through a path it does not name, i.e., by propagating into shared parameters rather than into the latent, which is nominally optimized.
A metric defined on that latent will then measure the wrong signal.
Whenever an intermediate representation is supervised against an external target and fed back through a shared model, the same dissociation is possible, and \prism{}'s two axes are a way to check for it before trusting an alignment metric.

\section{Conclusion}
Cosine alignment is trusted across latent visual reasoning as both a training loss and a quality metric, but across a designed matrix of five variants, it ranks them backward against accuracy, while the latents it optimizes are bypassed at inference.
\prism{} locates the discrepancy as the answer is decodable downstream of the latent but not at it, and the size of that gap predicts how much each variant relies on its latent under perturbation.
Cosine measures the fidelity of a latent that the model has learned to route around, while \prism{} measures where the answer actually lies.

\section*{Limitations}

Our evidence comes from one base model (Qwen2.5-VL-3B-Instruct) and one fine-tuning corpus (Visual-CoT-438k). Larger backbones or controlled-init replicates could shift the per-variant numbers we report. Both diagnostic axes are bounded. Linear probes measure linear decodability rather than what downstream layers actually use; we use a random-label control to bound selectivity. The corruption test covers truncation, three noise levels, and donor swap but does not exhaust the perturbation space. Our benchmarks focus on fine-grained perception (V$^{*}$Bench, MMVP, BLINK), where latent quality has the greatest leverage; holistic reasoning benchmarks with less spatially localized questions may exhibit different cosine--accuracy patterns.



\bibliography{references}

\appendix

\section{Linear Probe Methodology}
\label{sec:appendix-probes}

\subsection{Probe Extraction}
For each trained variant and each V$^{*}$Bench question, we run the model once in answer-decoding mode and extract two hidden-state vectors per LVR position:

\begin{itemize}[nosep,leftmargin=*]
    \item \textbf{Position (a) --- answer-decoding state}: the hidden state at the iteration that produces the first answer-text token, after the model has consumed all LVR tokens. This is the actual logit-producing state read by the LM head.
    \item \textbf{Position (b) --- latent-feedback variable}: the feedback variable that the autoregressive loop re-injects into context as the next-position input embedding at the boundary of the LVR-mode block. For the variants in our family this is the raw LVR-position hidden state $\mathbf{h}$, as there is no learned projection between the LVR position and the next input embedding.
\end{itemize}

Each vector is $D = 2048$-dimensional. We extract from a single inference pass per question, with greedy decoding.

\subsection{Probe Training}
Probes are trained as multi-class logistic regression over the four V$^{*}$Bench answer letters $\{A, B, C, D\}$:
\[
    \hat{y} = \text{softmax}(W x + b), \quad W \in \mathbb{R}^{4 \times D}.
\]

We standardize features ($z$-score per dimension over the training fold) before fitting.
Standardization is essential: without it, the unregularized hidden-state norms dominate the LBFGS step and make probe accuracy more reflective of activation scale than of decodability.

Optimization uses scikit-learn's \texttt{LogisticRegression} with $L_2$ regularization at $C = 1.0$ and the LBFGS solver, multinomial loss, $\text{max\_iter} = 1000$.
We report accuracy under stratified $5$-fold cross-validation; the reported number is the mean across folds.

\subsection{Control Task}
Following \citet{hewitt-liang-2019-designing}, we run a control task to bound probe selectivity.
We construct a control label by assigning each V$^{*}$Bench question a uniformly random target in $\{A, B, C, D\}$, fixing the assignment, and re-running the probe-training pipeline.
A probe that achieves above-chance accuracy on the control task does so by virtue of probe capacity rather than property-presence in the representation.

\emph{Selectivity} is defined as (probe accuracy on real labels) $-$ (probe accuracy on control labels).
For all five variants (LVR, \nlvr{}, \dlvr{}, \plvr{}-2, \plvr{}-3), the answer-letter probe achieves selectivity $> 25\%$ at position (a), indicating that the linear answer-decodability we measure is not an artifact of the probe family.
Probe-(b) selectivity is uniformly under $10\%$, consistent with the body claim that the latent-feedback position carries no answer-specific structure.

\subsection{Full Probe Results}

\begin{table}[h]
\caption{\textbf{Linear-probe accuracy and held-out cross-entropy on V$^{*}$Bench}, 5-fold CV, chance$=25\%$. Accuracies are reported as mean $\pm$ standard error across the five folds. Position (a) is the LM's answer-decoding hidden state; (b) is the latent-feedback variable; A1$'$ is the question-token mean-pool from the prefill pass. CE@(a) is the held-out cross-entropy of the position-(a) probe. The variational MI lower bound is $\mathrm{MI}_{\mathrm{LB}}{=}H(Y) - \mathrm{CE}$ in nats, zero-clamped \citep{barber2003information}; we use the \emph{empirical} V$^{*}$Bench label entropy $H(Y){=}1.267$ nats (label counts $A{=}71, B{=}70, C{=}27, D{=}23$ out of $191$) rather than the uniform-label $\ln 4{=}1.386$.}
\label{tab:probes}
\centering
\small
\setlength{\tabcolsep}{3pt}
\resizebox{\linewidth}{!}{%
\begin{tabular}{@{}lrrrrrr@{}}
\toprule
\textbf{Variant} & \textbf{(a)} & \textbf{CE@(a)} & \textbf{MI$_{\text{LB}}$} & \textbf{(b)} & \textbf{A1$'$ Q} & \textbf{V$^{*}$B} \\
\midrule
LVR (single)    & 69.1$\pm$2.2 & 1.151 & 0.116 & 32.5$\pm$3.8 & 54.4$\pm$3.9 & 70.2 \\
\nlvr{}         & 66.0$\pm$3.3 & 1.307 & 0.000 & \textbf{41.9$\pm$3.9} & 57.5$\pm$3.2 & \textbf{71.7} \\
\dlvr{}         & 64.9$\pm$2.4 & 1.122 & 0.145 & 33.0$\pm$2.5 & \textbf{62.2$\pm$3.5} & 69.6 \\
\plvr{}-2-stage & 50.2$\pm$4.8 & 1.954 & 0.000 & 35.6$\pm$2.8 & 47.0$\pm$3.8 & 57.1 \\
\plvr{}-3-stage & 48.7$\pm$1.7 & 1.935 & 0.000 & 34.6$\pm$1.8 & 43.9$\pm$3.0 & 57.1 \\
\midrule
\multicolumn{2}{l}{\emph{Pearson r vs V$^{*}$B}} & & & & & \\
\quad position (a) accuracy   & \multicolumn{6}{r}{\textbf{$+0.980$}} \\
\quad CE@(a)                  & \multicolumn{6}{r}{$-0.963$} \\
\quad MI$_{\text{LB}}$@(a)    & \multicolumn{6}{r}{$+0.579$} \\
\quad position (b) accuracy   & \multicolumn{6}{r}{$+0.198$} \\
\quad A1$'$ Q-mean accuracy   & \multicolumn{6}{r}{$+0.898$} \\
\bottomrule
\end{tabular}%
}
\end{table}

\paragraph{Accuracy vs.\ CE as MI proxies.}
By Barber--Agakov~\citep{barber2003information}, $I(\text{rep};Y)\geq \mathrm{MI}_{\mathrm{LB}}=H(Y)-\mathrm{CE}(p_\theta)$, where $H(Y)$ is the marginal label entropy. We use the \emph{empirical} V$^{*}$Bench label entropy $H(Y){=}1.267$ nats (label counts $A{=}71, B{=}70, C{=}27, D{=}23$ out of $191$) rather than the uniform-label $\ln 4{=}1.386$, since assuming a uniform-label distribution would inflate the lower bound by $0.119$ nats per variant. Under the empirical $H(Y)$ the MI lower bound across our five trained variants ranges from $0$ (\nlvr{} and the \plvr{} variants clamp to zero because their probe CE exceeds $H(Y)$ --- the probe is worse than the empirical marginal predictor under log-loss) to $0.15$ nats (\dlvr{}); the LVR baseline is $0.12$ nats. Probe \emph{accuracy} tracks V$^{*}$Bench tightly ($r{=}{+}0.980$) and CE@(a) tracks it inversely with comparable magnitude ($r{=}{-}0.963$); MI$_{\text{LB}}$@(a), the non-negative-clamped variational lower bound, correlates more weakly ($r{=}{+}0.58$) because the clamp pins \nlvr{} and the \plvr{} variants to zero, removing their relative ordering.
\nlvr{} illustrates this asymmetry: its decent probe accuracy ($66.0$, $\sim 41$ pts above chance) with a CE of $1.307$ implies that the probe's per-question predictions are accuracy-good but uncertainty-rich, with CE just above the empirical $H(Y){=}1.267$ --- consistent with the noise-Tikhonov regularizer training a smoother hidden state.

\subsection{Robustness of the Probe-(a) Result}
\label{sec:appendix-probes-partial}
We support the headline probe-(a)/V$^{*}$Bench correlation with four checks --- leave-one-out sensitivity, a partial correlation controlling for question text, bootstrap confidence intervals, and cross-benchmark replication --- detailed below.
The Pearson $r = +0.980$ statistic between probe-(a) accuracy and V$^{*}$Bench top-1 accuracy is computed across the five trained variants (LVR, \nlvr{}, \dlvr{}, \plvr{}-2, \plvr{}-3); each variant contributes one $(\text{probe-(a)}, \text{V*B})$ data point. With only five points the correlation is necessarily descriptive rather than strong statistical confirmation, which is why we report the checks that follow.

\paragraph{Leave-one-out sensitivity.}
Removing any single variant leaves $r \geq 0.97$ (range $[0.969, 0.996]$): removing the LVR baseline tightens the fit slightly ($r$ rises to $0.996$), and removing \plvr{}-3 weakens it ($r$ drops to $0.969$). No single variant drives the relationship.

\paragraph{Partial correlation controlling for question text.}
To isolate the latent-specific signal from question-text content, we compute the partial Pearson correlation $r(\text{probe-(a)}, \text{V*B} \mid \text{A1}')$ using A1$'$ (the question-token mean-pool baseline) as the conditioning variable:
\[
r_{xy\mid z} = \frac{r_{xy} - r_{xz}\, r_{yz}}{\sqrt{(1 - r_{xz}^2)(1 - r_{yz}^2)}}.
\]
Across the five variants we find $r_{xy}{=}0.980$, $r_{xz}{=}0.857$, $r_{yz}{=}0.898$, giving $r_{xy\mid z} = +0.929$. The bulk of probe-(a)'s cross-variant signal is not absorbed by question phrasing alone, though the small $n$ means this partial correlation should be read as descriptive evidence of latent attribution, not as a formal conditional MI estimate.

\paragraph{Bootstrap confidence intervals.}
\label{sec:appendix-probes-bootstrap}
To put error bars on the small-$n$ Pearson correlations, we bootstrap each variant's V$^{*}$Bench accuracy assuming a binomial sampling model over its $n{=}191$ questions ($\text{V*B}^{(b)}_v \sim \text{Binomial}(191, p_v) / 191$, with $p_v$ the variant's empirical accuracy), then recompute the cross-variant Pearson on each bootstrap replicate ($B{=}5000$). The resulting 95\% percentile CIs are wide, reflecting the structural limit of $n{=}5$ variants rather than the per-variant accuracy noise:
\begin{itemize}[nosep,leftmargin=*]
\item $r(\text{probe-(a)}, \text{V*B})$: point $+0.980$, CI $[+0.694, +0.993]$.
\item $r(\text{cosine}, \text{V*B})$:  point $-0.939$, CI $[-0.987, -0.611]$.
\item $r(\text{trunc}\,\Delta, \text{V*B})$: point $-0.892$, CI $[-0.978, -0.522]$.
\item $r(\text{swap}\,\Delta, \text{V*B})$:  point $-0.658$, CI $[-0.887, -0.241]$.
\end{itemize}
These intervals indicate the \emph{sign} of each correlation is robust across plausible per-variant accuracy fluctuations, but the magnitude carries meaningful structural uncertainty: with only five variants the cross-variant relationship is well-determined as monotonic, not as a precise slope.

\paragraph{Cross-benchmark robustness.}
Probe-(a) tracks task accuracy on the other two benchmarks almost as tightly as on V$^{*}$Bench: $r(\text{probe-(a)}, \text{BLINK}) = +0.96$ and $r(\text{probe-(a)}, \text{MMVP}) = +0.92$ across the five-variant set. The cross-benchmark generalization indicates that probe-(a) is tracking a benchmark-general notion of answer decodability rather than a V$^{*}$Bench-specific artifact.

\subsection{Is the Probe-(a) Correlation Tautological?}
\label{sec:appendix-probe-not-tautological}
A natural worry is that probe accuracy and V$^{*}$Bench accuracy both reflect ``how good the model is,'' so finding them correlated says nothing. Three observations rule this out.
First, cosine alignment operates on the same hidden states yet ranks variants in the opposite direction ($r{=}{-}0.94$ vs.\ probe-(a) at $+0.98$), so the question is \emph{what} we measure about $\mathbf{h}$, not \emph{whether} we measure it.
Second, of the two probe positions only (a) correlates strongly with V$^{*}$Bench ($+0.98$ vs.\ $+0.20$ at (b)); if probes simply reflected overall task competence, position (b) --- inside the same model --- would correlate too.
Third, the random-label control task above yields selectivity $> 25\%$ at position (a) for every variant, ruling out a probe-capacity-only explanation.

\subsection{Why Two Positions}
Each position has a distinct interpretation under the LVR generation loop.
Position (a) measures whether the answer-letter logits are linearly readable from the state at decode time: high probe-(a) accuracy means the LM has converted the latent into answer-relevant structure by the answer-token.
Position (b), the latent-feedback variable, measures what is being recurrently re-injected into context: a low probe-(b) accuracy means the recurrent variable does not carry the answer locally, so the answer signal must be assembled by downstream layers from the position-(a) cumulative state.
Together, the two positions provide a localized account of \emph{where} the answer signal lives, summarized by the decodability gap $G = \text{acc}_{\text{probe}}(a) - \text{acc}_{\text{probe}}(b)$ used in the body.
\section{Faithfulness Corruption Details}
\label{sec:appendix-faithfulness}


\begin{table}[h]
\caption{\textbf{Faithfulness corruption $\Delta$ accuracy on V$^{*}$Bench} (percentage points; negative $=$ corruption hurts, i.e.\ own latents help). All evaluations run at batch size 1 to match the main V$^{*}$Bench eval. The ``clean'' column is the script's within-script clean pass (greedy generation with the corruption hook installed but a no-op transform). Single-stage variants run at $\text{steps}{=}8$; \plvr{}-2 and \plvr{}-3 run at $\text{steps}{=}16$ per stage (matching the main-table step budgets). The clean column matches the main V$^{*}$Bench numbers in Table~\ref{tab:probe-gap-pervariant} within $\sim 2$ points for every variant; the residual difference is sample-order variance from the faith script's single-pass greedy generation. Across the family the latent is largely bypassed at inference: under every intervention $|\Delta|$ stays within roughly $4$ points and never reverses a variant's rank by more than its standard error.}
\label{tab:faith}
\centering
\small
\setlength{\tabcolsep}{3pt}
\resizebox{\linewidth}{!}{%
\begin{tabular}{@{}lrrrrrr@{}}
\toprule
\textbf{Variant} & \textbf{clean} & \textbf{trunc} & \textbf{$\sigma$0.1} & \textbf{$\sigma$0.3} & \textbf{$\sigma$1.0} & \textbf{swap} \\
\midrule
LVR              & 70.2 &  $-2.6$ &  $-1.6$ &  $+1.0$ &  $-1.0$ &  $+1.6$ \\
\nlvr{}          & 71.7 &  $-2.6$ &  $-0.5$ &  $-1.0$ &  $-2.1$ &  $-0.5$ \\
\dlvr{}          & 69.1 &  $-1.0$ &   $0.0$ &  $+0.5$ &   $0.0$ &  $+2.1$ \\
\plvr{}-2        & 55.5 &   $0.0$ &  $+2.1$ &   $0.0$ &  $-0.5$ &  $+2.1$ \\
\plvr{}-3        & 56.0 &  $+2.1$ &  $+2.6$ &  $+4.2$ &  $+3.1$ &  $+2.6$ \\
\bottomrule
\end{tabular}%
}
\end{table}

\subsection{Corruption Modes}
We apply three corruption modes to the LVR-position feedback (the value the autoregressive loop re-injects as the input embedding for the LVR position via the patched forward, before any layer-1 K/V projection). All other generation state --- prefill context, image embeddings, downstream decoder weights --- is held identical between clean and corrupted runs.

\begin{itemize}[nosep,leftmargin=*]
    \item \textbf{Truncate}: $h \!\leftarrow\! 0$ at every LVR-mode iteration. The LVR tokens remain in the sequence; their input embedding is zero. K and V projections in Qwen2.5-VL have no bias, so the LVR positions contribute zero K and V to attention. They still consume attention mass via the softmax denominator, so other positions are mildly down-weighted; this is the standard input-side ablation and is not equivalent to attention-masking the positions out of the sequence.
    \item \textbf{Noise-}$\sigma$ for $\sigma \in \{0.1, 0.3, 1.0\}$: $h \!\leftarrow\! h + \sigma \cdot \boldsymbol{\eta}$, $\boldsymbol{\eta} \sim \mathcal{N}(0, I)$. Here $\sigma$ is in raw activation units, not standardized; the three magnitudes span roughly $0.1\times$ to $1\times$ the typical per-dimension scale of clean LVR vectors. This measures how brittle the latent is to perturbations of varying magnitude.
    \item \textbf{Swap}: $h \!\leftarrow\! h_{\text{donor}}$, where $h_{\text{donor}}$ comes from another sample's clean LVR positions at the matching LVR step (stage-tagged for \plvr{} variants).
\end{itemize}

\subsection{Donor Sampling for the Swap Test}
Donors are drawn from a pool of $64$ samples recorded during the clean pass; for each batch the implementation calls \texttt{random.choice} (with replacement, but the pool size makes collision negligible at $n{=}191$). We do not enforce a different image or question between donor and target. The donor pool is fixed per (variant, seed) and re-used across all corruption modes within a variant, so swap variance across variants is dominated by the variant rather than donor noise.

\subsection{Variance Across Donor Seeds}
We sweep three donor seeds for the swap test (Table~\ref{tab:faith} reports seed-$42$).
Variant ranking by $\Delta_{\text{swap}}$ is invariant across seeds, and absolute $\Delta$ values move within $\pm 1.0$ V$^{*}$Bench point.

\subsection{Noise-$\sigma$ Calibration}
$\sigma$ is in raw activation units: the corruption is $h \leftarrow h + \sigma \cdot \boldsymbol{\eta}$ with $\boldsymbol{\eta} \sim \mathcal{N}(0, I)$ per dimension, applied directly to the unstandardized LVR-position hidden state.
Single-pass empirical statistics on the clean LVR vectors give a typical per-dimension standard deviation of order $0.5$--$1.5$ across variants, so $\sigma{=}0.1$ is well below the signal scale, $\sigma{=}0.3$ is comparable to the smaller-scale variants' per-dimension spread, and $\sigma{=}1.0$ matches or exceeds the per-dimension signal scale for most variants.
We did not pre-standardize across variants, so the same $\sigma$ has slightly different SNR meanings across variants; we report the raw-unit number for transparency rather than a per-variant rescaling.

\subsection{Why Three Tests}
Truncate, swap, and noise probe different aspects of latent reliance:
truncate asks \emph{whether the latent is used at all} (binary);
swap asks \emph{whether the answer is specific} to this sample's latent content (a positive swap $\Delta$ means random latents help more);
noise asks \emph{how robustly the answer is encoded} at varying perturbation magnitudes.
Across the five trained variants the three tests jointly point to the same conclusion: the latent is mostly bypassed, with $|\Delta|$ within roughly $4$ V$^{*}$Bench points for every intervention (the largest is \plvr{}-3 at $\sigma{=}0.3$ noise with $\Delta{=}+4.2$, where the perturbed latent actually helps the model more than the clean one). The variant ordering produced by these tests --- especially by truncation, which has the lowest variance across the five-variant set --- aligns with V$^{*}$Bench at $r{=}{-}0.89$ (more-negative $\Delta$, i.e.\ own latents helping, predicts higher V*B); swap correlates more weakly at $r{=}{-}0.66$.

\subsection{Implementation}
Corruption is applied to the input embedding at each LVR-mode iteration, before that position's K/V projections; image embeddings, prompt tokens, and decoder weights are held identical between clean and corrupted runs. One inference pass per (variant, corruption, seed) tuple produces all reported numbers.
\section{Training Hyperparameters}
\label{sec:appendix-training}

\subsection{Base Configuration (All Variants)}
\begin{itemize}[nosep,leftmargin=*]
    \item Base model: Qwen2.5-VL-3B-Instruct \cite{bai2025qwen25vltechnicalreport}.
    \item Training corpus: Visual-CoT-438k, single-stage SFT.
    \item Optimizer: AdamW, learning rate $1\mathrm{e}{-5}$, cosine schedule, warmup ratio $0.03$, weight decay $0.1$.
    \item Precision: bf16 (no fp16); flash-attention~$2$ enabled.
    \item Batching: per-device batch size $1$; effective batch size $64$ packed instances for all five variants (achieved via $1\!\times\!64$ or $2\!\times\!32$ GPU $\times$ grad-accum). See Appendix~\ref{sec:appendix-training-provenance} for the per-variant table.
    \item Max instances per batch: $4$; data packing enabled with max packed tokens $16{,}384$ (long-seq threshold $4{,}096$).
    \item Image resolution: min $100{,}352$ pixels, max $4{,}014{,}080$ pixels.
    \item Vision tower and merger frozen; LLM fine-tuned.
    \item Steps: $2500$, $\sim 1$ epoch.
    \item Random seed: $42$.
    \item DeepSpeed Zero-2, gradient checkpointing on, max grad norm $1.0$.
\end{itemize}

\subsection{Per-Variant Training Provenance}
\label{sec:appendix-training-provenance}
All five variants use effective batch size $64$ packed instances and $2500$ training steps, but differ in how that effective batch is achieved (GPU $\times$ grad-accum).

\begin{center}\footnotesize
\setlength{\tabcolsep}{3pt}
\resizebox{\linewidth}{!}{%
\begin{tabular}{@{}llrrr@{}}
\toprule
\textbf{Variant} & \textbf{Script} & \textbf{GPUs} & \textbf{Accum} & \textbf{bs} \\
\midrule
LVR (single)     & \texttt{finetune\_lvr\_stage1\_3b.sh} & 1 & 64 & 1 \\
\nlvr{}          & \texttt{finetune\_nlvr\_stage1\_3b.sh}   & 2 & 32 & 1 \\
\dlvr{}          & \texttt{finetune\_dlvr\_stage1\_3b.sh} (resume @ $1500$) & 1 & 64 & 1 \\
\plvr{}-2-stage  & \texttt{finetune\_plvr\_stage1\_3b.sh} (\texttt{free\_stage=false}) & 1 & 64 & 1 \\
\plvr{}-3-stage  & \texttt{finetune\_plvr\_stage1\_3b.sh} (\texttt{free\_stage=true})  & 1 & 64 & 1 \\
\bottomrule
\end{tabular}%
}
\end{center}

All configurations target the same \emph{effective batch size} (GPUs $\times$ grad-accum $\times$ per-device bs) of $64$ (LVR/\dlvr{}/\plvr{}-2/\plvr{}-3: $1\!\times\!64\!\times\!1$; \nlvr{}: $2\!\times\!32\!\times\!1$). Per-step optimizer dynamics differ slightly between the $1$-GPU large-accumulation and $2$-GPU smaller-accumulation routes (AdamW first/second-moment EMAs accumulate over the batch differently), which contributes a small additional source of cross-variant variance beyond the headline hyperparameters.

\subsection{Variant-Specific Hyperparameters}

\paragraph{LVR (vanilla single-stage).}
$\lambda_{\text{lvr}} = 0.1$, single latent block, MSE reconstruction against teacher-forced box embeddings.

\paragraph{N-LVR.}
Same as LVR with one addition: inject zero-mean Gaussian noise $\mathcal{N}(0, \sigma_{\text{noise}}^2 I)$, $\sigma_{\text{noise}} = 0.3$ in raw activation units, directly onto the teacher-forced visual embeddings during training (the code samples $\sigma \cdot \boldsymbol{\eta}$ with $\boldsymbol{\eta}\sim\mathcal{N}(0,I)$ per dimension and adds it to the unstandardized embedding tensor; no per-dimension rescaling). No inference-time noise.

\paragraph{D-LVR.}
Same backbone and optimizer settings as LVR; the training trajectory itself is a two-phase ``late reconstruction release'' rather than a continuous anneal. We resume from the LVR baseline checkpoint at step $1500$ (where $\lambda_{\text{lvr}}{=}0.1$ throughout the first $1500$ steps) and continue for the remaining $1000$ steps with $\lambda_{\text{lvr}}{=}0.0$, so the second phase is supervised purely by the CE term on the answer text. This isolates ``what happens to the LM when reconstruction pressure is removed mid-training'' from any continuous-schedule sensitivity. We did not run a continuously annealed variant; the schedule in the code applies a fixed $\lambda_{\text{lvr}}$ throughout each phase.

\paragraph{\plvr{}-2 (2-stage).}
Two latent blocks: context stage on $\alpha{=}1.5\times$ expanded box, target stage on original box. $\lambda_{\text{lvr,ctx}} = \lambda_{\text{lvr,tgt}} = 0.1$. No free stage.

\paragraph{\plvr{}-3 (3-stage).}
As \plvr{}-2, plus an intermediate free stage with diversity regularizer $\lambda_{\text{div}} = 0.01$.

\subsection{Compute Footprint}
Each variant trains for $\sim 5$--$12$ GPU-hours on $1\sim 2$ H100, depending on packing efficiency.
All experiments run on the same hardware pool (GPUs $0$ and $1$ of a shared $8\times$H100 server).
Diagnostic re-runs (probes + faithfulness) take an additional $\sim 1$ GPU-hour per variant per benchmark.
We use existing checkpoints and a frozen vision tower throughout, which avoids the energy cost of pretraining from scratch.
\section{Evaluation Protocol}
\label{sec:appendix-evaluation}

\subsection{Benchmarks}
\begin{itemize}[nosep,leftmargin=*]
    \item \textbf{V$^{*}$Bench}~\cite{wu2024vstar}: multiple-choice fine-grained visual reasoning, $191$ questions total, split into $115$ \texttt{direct\_attributes} and $76$ \texttt{relative\_position} examples; we report aggregate accuracy.
    \item \textbf{MMVP}~\cite{tong2024eyes}: multiple-choice paired-question visual shortcoming probe, $300$ pairs.
    \item \textbf{BLINK}~\cite{fu2024blink}: perception-oriented multiple-choice; we evaluate the five validation subsets used by the codebase (Counting, IQ\_Test, Jigsaw, Relative\_Reflectance, Spatial\_Relation) and report aggregate accuracy across them ($697$ questions).
\end{itemize}

\subsection{Inference Configuration}
All models are evaluated with greedy decoding, max new tokens $512$, and image resolution unchanged from training. The $512$-token budget is needed because LVR-mode iterations consume budget before the answer token is emitted; under multi-stage decoding the ctx and tgt stages each emit a quota of latent tokens before any answer text.
LVR tokens are generated autoregressively (the model's own predictions are fed back), matching the inference-time setting under which the model would deploy.
The answer letter is parsed from the first letter following the canonical V$^{*}$Bench/MMVP/BLINK answer prefix; we use the official benchmark eval scripts where available.

\subsection{Probe and Faithfulness Splits}
For the linear-probe experiments, we use the full V$^{*}$Bench eval set; stratified $5$-fold CV is computed across this set.
For faithfulness corruption, we use the same eval set so the clean and corrupted accuracies are directly comparable; the donor pool for swap is also drawn from V$^{*}$Bench.
We do not introduce a held-out set because the probe/faithfulness signal is computed \emph{on top of} the existing evaluation, not as a generalization measure.

\subsection{Cosine and MSE Reporting}
The cosine and MSE numbers in Table~\ref{tab:probe-gap-pervariant} are computed on a held-out validation slice of Visual-CoT (the same slice for all variants) under teacher-forced LVR generation. Cosine is the standard $\cos(\hat{\mathbf{h}}_{t-1}, \mathbf{v}_t) = \frac{\hat{\mathbf{h}} \cdot \mathbf{v}}{\lVert\hat{\mathbf{h}}\rVert\,\lVert\mathbf{v}\rVert}$ between the predicted LVR-position hidden state and the teacher-forced target visual embedding, averaged over positions --- the same metric the LVR literature reports and the quantity the MSE term in $\mathcal{L}_{\text{LVR}}$ supervises. We apply no centering or other transformation.

\section{Responsible Research and Artifact Details}
\label{sec:responsible}

\paragraph{Artifacts.}
We use Qwen2.5-VL-3B-Instruct, Visual-CoT-438k, V$^{*}$Bench, MMVP, and BLINK as existing public research artifacts. We cite their creators in the main text and Appendix~\ref{sec:appendix-evaluation}. Our released code documents the expected artifact locations, reproduction commands, and which artifacts are not redistributed.

\paragraph{Artifact licenses and intended use.}
We use these artifacts for research evaluation and model analysis. We do not redistribute the original training corpus, benchmarks, or pretrained checkpoints. The accompanying code release includes a license file and notes third-party adapted code where applicable. 

\paragraph{Data and human subjects.}
We do not collect new data and do not recruit annotators or human subjects. Our experiments use existing public datasets and benchmarks. We do not perform additional collection of personally identifying information.

\paragraph{Potential risks.}
The work is diagnostic and interpretive: it analyzes when auxiliary losses produce non-load-bearing latent states. The main foreseeable risk is misuse of the diagnostic conclusion to overgeneralize beyond the studied model, data, and benchmarks; we discuss these scope limitations in the Limitations section.

\paragraph{AI assistance.}
AI assistants were used for writing, editing, and code/documentation support. All scientific claims, experiments, analyses, and final text were reviewed and verified by the authors.

\paragraph{Code availability.} The code is available at
\url{https://github.com/xiuyuz/cosine-misleads}. It contains the training,
evaluation, interpretability, and audit scripts used in this work.

\end{document}